\documentclass[runningheads]{llncs}
\usepackage{graphicx}
\usepackage{color}
\usepackage{amssymb}
\usepackage{booktabs}
\begin{document}
\title{Learning to Detect Collisions for Continuum Manipulators without a Prior Model}
\titlerunning{Learning to Detect Collisions for Continuum Manipulators}
%\author{}
%\institute{}
%\author{Shahriar Sefati\inst{1}\orcidID{0000-1111-2222-3333} \and
%Shahin Sefati\inst{2}\orcidID{1111-2222-3333-4444} \and
%Iulian Iordachita\inst{1}\orcidID{2222--3333-4444-5555} \and Russell H. Taylor\inst{1}\orcidID{2222--3333-4444-5555} \and Mehran Armand\inst{1}\orcidID{2222--3333-4444-5555}}

\author{Shahriar Sefati\inst{1} \and
Shahin Sefati\inst{2} \and
Iulian Iordachita\inst{1} \and \\ Russell H. Taylor\inst{1} \and Mehran Armand\inst{1,3}}
%index{Sefati, Shahriar}
%index{Sefati, Shahin}
%index{Iordachita, Iulian}
%index{Taylor, Russell}
%index{Armand, Mehran}
%
\authorrunning{S. Sefati et al.}
\institute{Laboratory for Computational Sensing and Robotics, Johns Hopkins University, Baltimore MD 21218, USA \email{\{sefati,iordachita,rht\}@jhu.edu}\and
Comcast Applied AI Research, Comcast, Washington D.C. 20005, USA
\email{shahin\_sefati@comcast.com}\\ \and
Johns Hopkins University Applied Physics Laboratory, Laurel 20723, USA\\
\email {mehran.armand@jhuapl.edu} }
\maketitle              % typeset the header of the contribution
\begin{abstract}
Due to their flexibility, dexterity, and compact size, Continuum Manipulators (CMs) can enhance minimally invasive interventions. In these procedures, the CM may be operated in proximity of sensitive organs; therefore, requiring accurate and appropriate feedback when colliding with their surroundings. Conventional CM collision detection algorithms rely on a combination of exact CM constrained kinematics model, geometrical assumptions such as constant curvature behavior, \textit{a priori} knowledge of the environmental constraint geometry, and/or additional sensors to scan the environment or sense contacts. In this paper, we propose a data-driven machine learning approach using only the available sensory information, without requiring any prior geometrical assumptions, model of the CM or the surrounding environment. The proposed algorithm is implemented and evaluated on a non-constant curvature CM, equipped with Fiber Bragg Grating (FBG) optical sensors for shape sensing purposes. Results demonstrate successful detection of collisions in constrained environments with soft and hard obstacles with unknown stiffness and location.

\end{abstract}
\section{Introduction}
% paragraph about use of continuum manipulators in surgery
% advantages over rigid-link robots, challenges in sensing, control and detecting collision
Compared to conventional rigid-link robots, CMs exhibit higher dexterity, flexibility, compliance, and conformity to confined spaces, making them suitable for minimally invasive interventions. Examples of the use of CMs in medical applications include, but not limited to neurosurgery, otolaryngology, cardiac, vascular, and abdominal interventions~\cite{burgner2015continuum}. In these medical interventions, the CM may be used for steering in curvilinear pathways, manipulating tissue, or merely as a flexible pathway for navigation of flexible instruments to a desired surgical site, all of which accentuating the necessity of detecting CM collision or contact with bone, tissue, or organs.

% Collision detection in continuum manipulators 
% bridge toward motivation for model-independent methods

Detecting CM collisions with the environment is of great importance and has been studied in the literature. In~\cite{bajo2012kinematics}, CM collisions were detected by monitoring the distance between expected and the actual instantaneous screw axis of motion, measured from the unconstrained CM kinematics model and electromagnetic sensory information, respectively. In~\cite{li2012exact}, collisions were detected by obtaining the exact model of a continuum manipulator featuring multiple constant-curvature sections and a modeled object using polygonal mesh. In~\cite{bajo2010finding}, two model-dependent approaches were presented: 1) using the deviation of joint forces from the nominal CM model, and 2) using a modified CM kinematics model to detect contacts. These studies, in addition to sensory information, required exact modeling of the CM or objects, relying on geometrical assumptions and properties specific to the choice of continuum manipulator.

% Advances of machine and deep learning in various areas, and robotics especially, in control and recently in sensing
This study aims at enhancing safety during teleoperation or autonomous control of CMs in proximity of sensitive organs in confined spaces. We propose a data-driven machine learning algorithm that solely relies on data from any already-available sensor, independent of the CM kinematics model, without any prior assumption or knowledge regarding the geometry of the CM or its surrounding environment. The key idea behind our proposed method is to define the problem of collision detection as a classification problem, with sensory information as the input, and occurrence/no-occurrence of CM collision with the environment as the output classes. A machine learning model is trained preoperatively on the sensory data from the CM, and is then used intraoperatively to detect collisions with the unknown environment. This information can then be conveyed to the surgeon as audio or haptic feedback to safeguard against potential trauma and damage to sensitive organs.

% General idea of pre-operative training and intra-operative collision detection, maybe talk about 
% FBG
Different sensing methods such as electromagnetic tracking and intraoperative imaging have been used for sensing in CMs~\cite{shi2017shape}. In recent years, FBG optical sensors have offered advantages over other sensing methods such as not requiring a direct line of sight, high streaming rate (up to $1$ KHz), and minimal effects on compliance and compactness. FBG sensors have been used for shape, force, and torsion sensing in applications with CMs, biopsy needles, catheters, and other flexible medical instruments~\cite{jackle2019shape,khan2019multi,lai2018distal,rahman2019modular,sefati2016fbg,sefati2018effect,xu2016curvature}. To this end, we implement our proposed method on a CM previously developed for MIS applications, equipped with FBG sensors for shape estimation~\cite{sefati2019fbg,sefati2018fbg,sefati2017highly}. Using the proposed method, the FBG sensor will serve as a dual-purpose simultaneous shape sensor and collision detector, preserving CM small size from additional sensors.

The contributions of this work include 1) the development of a supervised machine learning algorithm for data-driven CM collision detection with audio feedback, without requiring additional sensors, a prior model of the CM, obstacle location, or environment properties, 2) training and tuning of the hyper-parameters for the machine learning algorithm, and 3) verification of the algorithm by performing a series of experiments involving CM collision with hard and soft objects with unknown properties and locations.

%Tailored towards the application and the specific CM, FBG sensors with a variaty of design, configuration, and integration with the CM have been explored in the literature []. 

\begin{figure}[b!]
\centering
\includegraphics[width=0.5\textwidth]{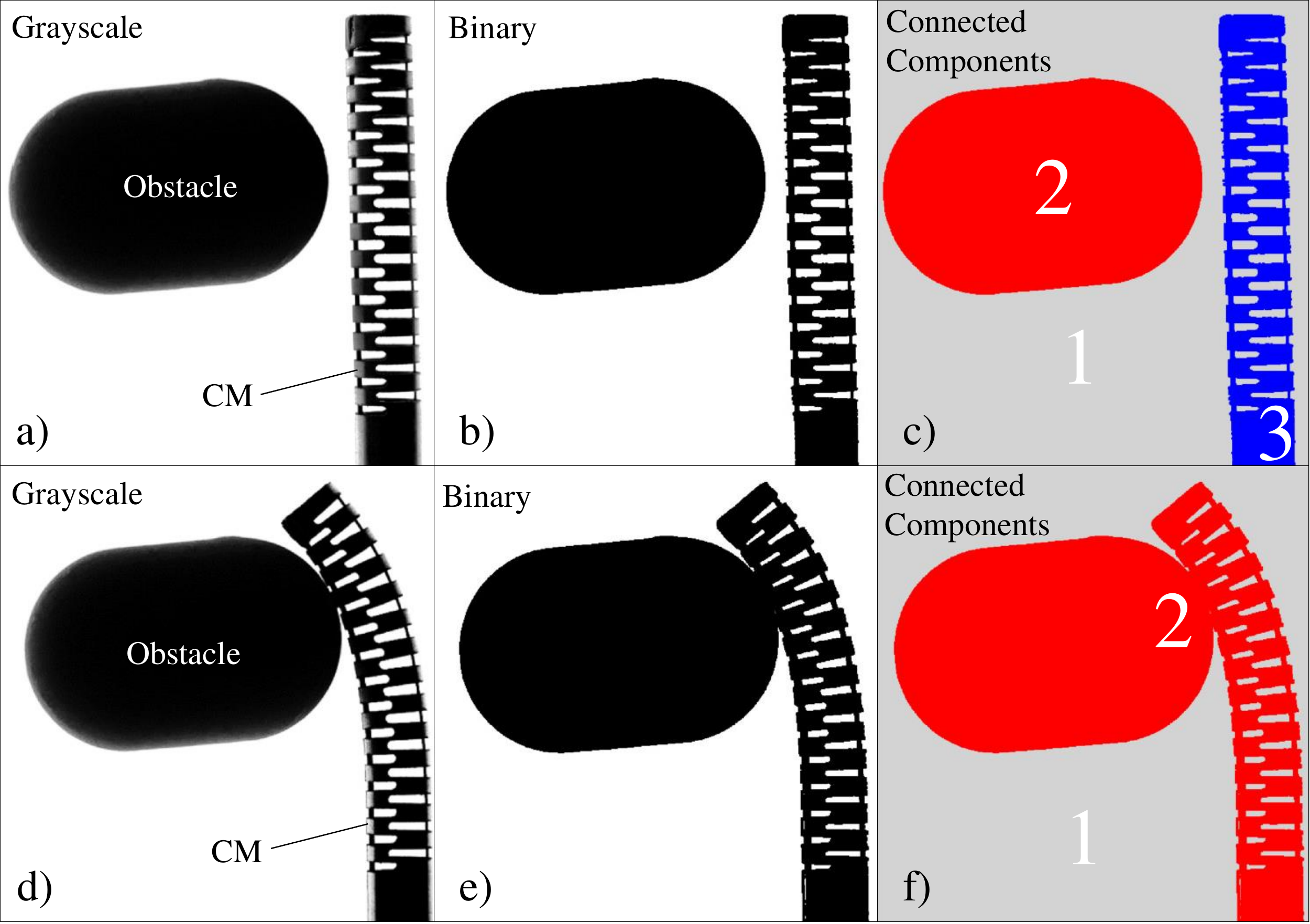}
\caption{Image collision labeling based on number of connected components. Top and bottom rows demonstrate no collision and collision labels, respectively.} \label{fig:vision-based-alg}
\end{figure}

\section{Method} \label{Method}
We define the problem of collision detection as a supervised classification problem with two classes: collision, and no collision. The input to this method is the sensory data obtained from the CM and the output is the corresponding class of collision. Similar to other supervised learning algorithms, the proposed method consists of an offline dataset creation step during which, the sensory data is labeled with correct collision class. A supervised machine learning model is trained on the collected dataset to learn the nonlinear mapping from sensory data to the appropriate class of collision. The trained model is optimized by tuning the hyper-parameters via a K-fold cross validation by splitting the data to training and validation sets. Performance of the tuned model is then evaluated on unseen test data from CM collisions with obstacles with different stiffness and properties (hard and soft), placed at random unknown locations relative to the CM.

\subsection{Dataset Creation} \label{DatasetCreation}
To create an appropriate dataset, a vision-based algorithm based on a connected components labeling algorithm~\cite{he2009fast} is used to segment preoperative images captured via an overhead camera looking at the CM and the surrounding obstacles. The images are first converted to binary format by applying appropriate thresholds. An erosion morphological operation~\cite{comer1999morphological}, followed up with a dilation, both with small kernel sizes are applied to the binary image to remove potential background noise and ensure robust connected region detection. The connected components labeling algorithm then segments the binary image to distinguish between the background, the CM, and obstacles present in the scene. A particular sensor sample is labeled with the collision class, if the CM and the obstacle form a connected region in the corresponding image frame. Fig.~\ref{fig:vision-based-alg} indicates the segmented regions in collision and no collision instances during the training phase. 

To follow a simple and easy-to-repeat dataset creation procedure, a random-sized oval-shaped obstacle was 3D-printed and placed in five random locations in the CM surrounding volume. The CM was then actuated to collide with the obstacles, while capturing synchronized sensory data and overhead camera images. The captured images were then segmented using the vision-based algorithm to generate the corresponding collision labels.

\begin{figure} [b!]
\includegraphics[width=\textwidth]{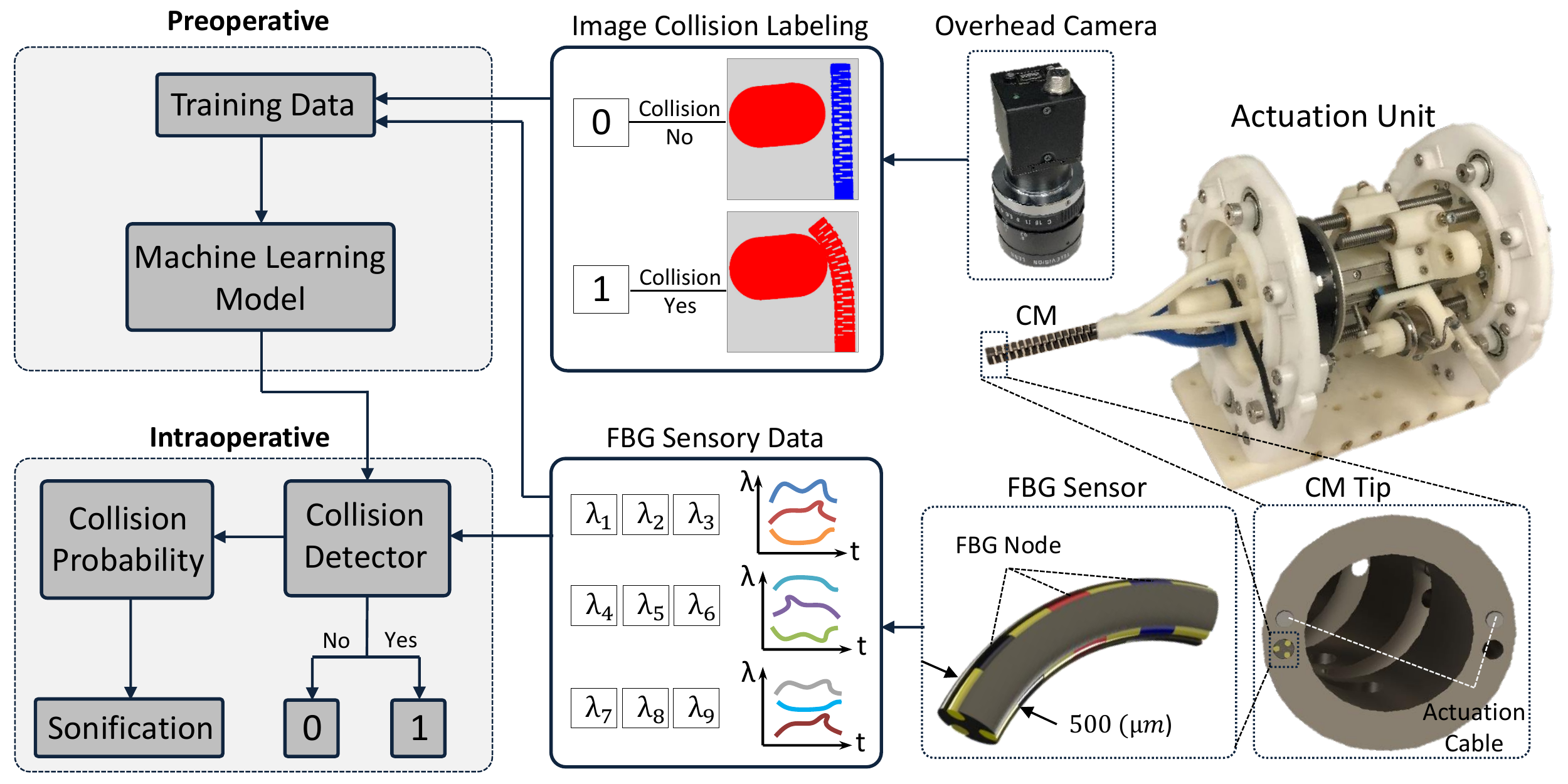}
\caption{Experimental setup and the proposed framework: preoperative phase involving model training using sensory data and camera images, and intraoperative phase for collision detection and sonification using only sensory data.} \label{fig:setup}
\end{figure}

\subsection{Data-driven Collision Detection Framework}

Most conventional CM collision detection algorithms are physics- and geometric-based models. These approaches may lead to very complex models that do not accurately detect collisions. This might be in part due to the lack of knowledge of reliable physical models of the components, and/or due to lack of knowledge of the topology of interacting components. To this end, we propose a data-driven approach to directly identify a collision detection model for CMs based on empirical data. We formulate the collision detection problem as a machine learning classification problem. In particular, we use a gradient boosting classifier to learn and build a collision detection model for CMs. Gradient boosting trees, similar to most other ensemble models, are less likely to overfit to training data which are suitable for generalization of the model to different obstacles placed at unseen locations. Fig.~\ref{fig:setup} demonstrates the proposed framework.

\subsubsection{Gradient Boosting Classifier}  \label{GradientBoostingTree}
Gradient boosting is a powerful machine learning technique for classification and regression problems. Gradient Boosting builds a model in a sequential way and combines an ensemble of relatively weak simple models (base learners) \cite{friedman2001greedy}. Let $\{x_k\}_{k=1}^N$ be a sequence of observations, where $x_k \in \mathbb{R}^n$ ($n$ is the number of FBG sensory data at frame $k$) represents the observation at frame $k$. Let $y_k$ be their corresponding labels ($1$ collision, $0$ no collision). Given the training data, we train a Gradient Boosting Classifier to detect collision. The classifier scores can be used as probability of predicting collection. 

%To prevent overfitting to training data, we apply regularization techniques such as shrinkage and subsmapling (see \ref{Results}).

\subsection{Real-time Sonification} \label{AudioFeedback}
Visual feedback is one of the most common means of conveying information to the surgeon during surgery. However, augmentation of this feedback method with many sources of information is prone to the risk of missing crucial information. %Particularly, during teleoperation and control of CMs in confined spaces, the visual feedback may be compromised by occlusion or miss-interpretation due to cluttered background; increasing the risk of damaging sensitive organs or tissues. 
As a workaround to these challenges, we convey the information associated with the collision detection via sonification of the probability of predicting collision (classifier score).

\section{Experiments} \label{Experiments}

\subsection{Experiment Design} \label{Experiment-Design}
To create the dataset explained in section \ref{DatasetCreation}, we 3D printed an oval-shaped obstacle from plastic ABS material, placed it at five random locations around the CM body, and collected the sensory data and overhead images, which we refer to as \textit{offline} dataset. This dataset was the baseline for training and tuning the collision detector. Additionally, to evaluate the trained collision detector's performance and robustness on unseen data, we designed experiments that involved CM collisions with objects with different shapes and stiffness, such as soft gelatin, sponge foam, and also plastic ABS, but placed at locations different from the ones in the \textit{offline} dataset, and also hand collisions at various points along the CM body.

\subsection{Experimental Setup and Software} \label{Experiment-Setup}
Fig.~\ref{fig:setup} demonstrates the used experimental setup. It consisted a cable-driven CM previously developed for MIS, constructed from a 35 mm-long nitinol (NiTi) tube with outside and inner diameters of 6 mm and 4 mm, respectively, suitable for passing flexible instruments through the CM. The CM was equipped with a fiber optic sensor consisted of 3 FBG fibers (each with 3 FBG nodes) attached to 0.5 mm outside diameter NiTi wire. The CM cables were actuated by two DC motors (RE16, Maxon Motor Inc. Switzerland) with velocity or position control of the actuation cables. A Flea2 1394b (FLIR Integrated Imaging Solutions Inc.) camera was used to capture the overhead images for dataset creation. Model training and testing were programmed in Python using Scikit-learn open source library~\cite{scikit-learn}, and UDP communication transferred the trained model's output data to Max/MSP (Cycling '74, San Francisco, USA) for sonification. FBG data and camera images are streamed and recorded at $100$ Hz and $30$ Hz, respectively.

\section{Results} \label{Results}
\begin{figure}
\includegraphics[width=\textwidth]{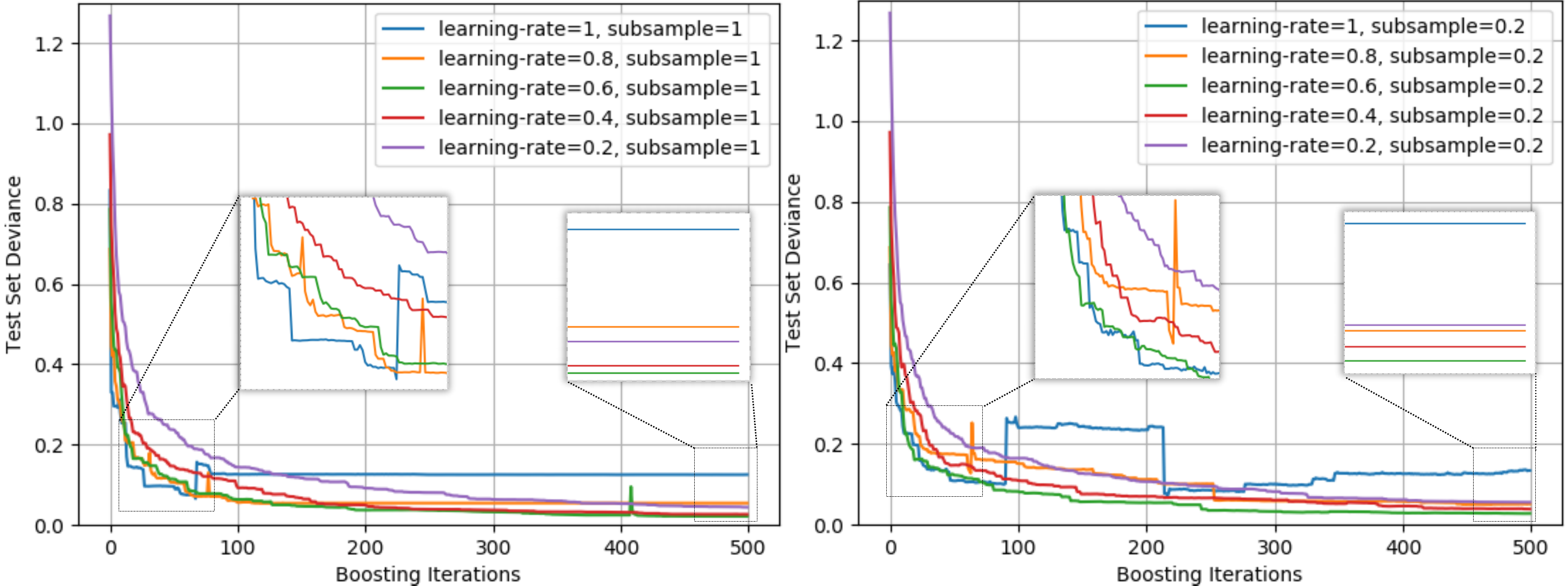}
\centering
\caption{Regularization via shrinkage and subsampling to avoid overfitting and enhance generalization. X and Y axis correspond to boosting iterations and loss on test data, respectively.} \label{fig:learning-rate}
\end{figure}

\begin{figure}
\includegraphics[width=\textwidth]{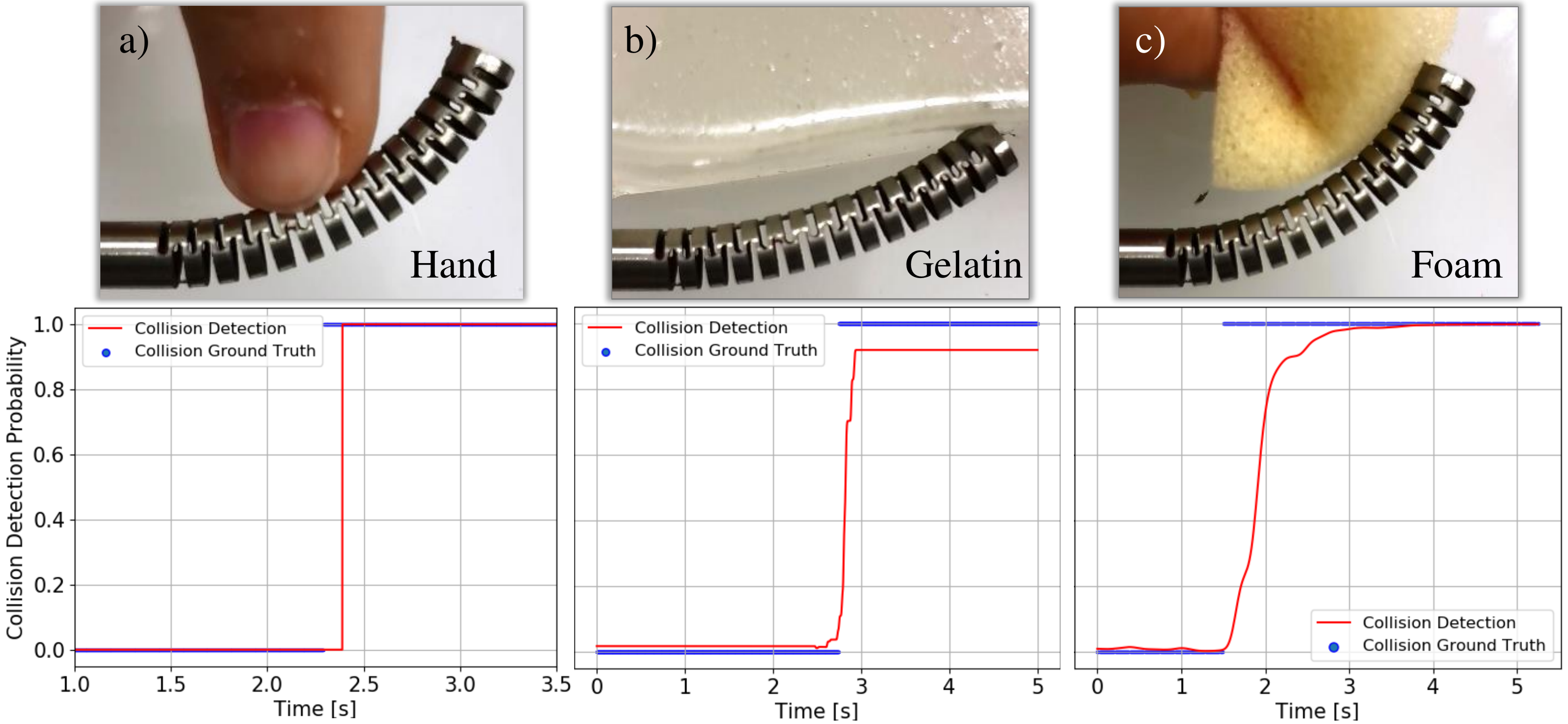}
\caption{Real-time collision detection on unseen data (unknown obstacle stiffness and location). CM collides with a) hand b) soft gelatin phantom, and c) soft sponge foam.} \label{fig:collision-result}
\end{figure}

To establish a baseline for optimal collision detection performance, the boosting tree hyper-parameters such as learning rate, number of max features, max depth, and boosting iterations (number of estimators) were tuned by running a randomized k-fold cross-validation on the \textit{offline} dataset. To avoid overfitting to the particular \textit{offline} dataset, and to enhance generalization to future unseen observations, we explored different regularization techniques such as shrinkage (learning rate $< 1$), stochastic gradient boosting (subsample $< 1$), and variation on maximum number of features (using all or $log2$ of the features). An initial a k-fold ($k=4$) validation on different number of estimators ($\{100, 250, 500\}$) and max depths ($\{3, 5, 7\}$) yielded optimal performance with these parameters set to $500$ and $3$, respectively. Fig.~\ref{fig:learning-rate} demonstrates the performance results on the \textit{offline} dataset with $57000$ samples for five experiments, over $500$ boosting iterations, with max depth of $3$, and with different combinations of learning rates ($\{1, 0.8, 0.6, 0.4, 0.2\}$) and subsampling values ($\{1, 0.2\}$). It is observed that shrinkage yields improvements in model's generalization ability over gradient boosting without shrinking (left figure). Additionally, subsampling can increase performance when combined with shrinkage (right figure and Table~\ref{table:results}). Table~\ref{table:results} summarizes boosting tree performance results for various combinations of learning rates, subsampling values, number of maximum features, and training times. Comparing the results, the optimal hyper-parameters (shown in bold font in table~\ref{table:results}) are chosen to maximize the k-fold mean accuracy and minimize the standard deviation of performance among all combinations of training and testing sets in a k-fold split. The performance of this optimized model is then evaluated on unseen data in real-time collisions with hard and soft objects different from the ones in the \textit{offline} dataset, placed at new locations around the CM. Fig.~\ref{fig:collision-result} indicates successful CM collision detection with hand, gelatin, and foam. The collision probabilities from the gradient boosting collision detector with $9$ FBG sensory data as input are shown and compared to ground truth collisions from the overhead camera. Please refer to the supplementary material for additional scenarios of CM collision with other materials at unknown locations with different properties. Of note, the time needed for \textit{offline} dataset collection is within a few minutes, depending on the number of locations for obstacle placement during the training phase.

\begin{table}
\caption{K-fold ($k=4$) cross validation results for hyper-parameter tuning.}\label{table:results}
\centering
\begin{tabular}{c|c|c|c|c|c|c|c|c|c|c|c|c}
\hline
Learning Rate & \multicolumn{4}{c|}{0.2} & \multicolumn{4}{c|}{\textbf{0.6}} & \multicolumn{4}{c}{1} \\ \hline
Max Features & \multicolumn{2}{c|}{all} & \multicolumn{2}{c|}{log2} & \multicolumn{2}{c|}{all} & \multicolumn{2}{c|}{\textbf{log2}} & \multicolumn{2}{c|}{all} & \multicolumn{2}{c}{log2} \\ \hline
Sub-Sample & 1.0 & 0.2 & 1.0 & 0.2 & 1.0 & 0.2 & \textbf{1.0} & 0.2 & 1.0 & 0.2 & 1.0 & 0.2 \\ \cline{1-1}
Mean Accuracy (\%) & 97.8 & 97.8 & 97.7 & 97.5 & 97.7 & 98.2 & \textbf{98.6} & 98.0 & 97.0 & 93.5 & 82.1 & 97.9 \\ \cline{1-1}
Std. Deviation (\%) & 0.05 & 0.07 & 0.07 & 0.03 & 1.48 & 0.14 & \textbf{0.04} & 0.20 & 2.06 & 6.95 & 18.0 & 0.23 \\ \cline{1-1}
Training Time (s) & 108.0 & 65.3 & 47.5 & 40.4 & 94.7 & 64.2 & \textbf{46.1} & 40.0 & 77.9 & 64.5 & 41.9 & 37.9 \\ \hline
\end{tabular}
\end{table}

\section{Discussion and Conclusion} \label{Conclusion}
We proposed a data-driven machine learning approach to collision detection in CMs without prior models, using only the available sensory data such as FBG. The proposed framework consisted of a preoperative offline training and tuning step via k-fold cross validation on a dataset created with a 3D-printed object placed at five locations around the CM, and an intraoperative online collision detection step, as well as sonification, during which only the FBG data was fed to the trained model. Results demonstrated successful detection of collisions on unseen data in constrained environments with soft and hard obstacles of different properties and unknown location. It should be noted that other types of sensors (e.g. Electromagnetic tracking) could be substituted for FBGs, since the algorithm only relies on the raw sensory data. Additionally, even though the CM used in the experiments is restricted to planar bending, the FBG sensor is capable of detecting 3D motions (since using 3 fibers), therefore, the method could potentially be extended to 3D manipulators. Such collision detection algorithm could enhance safety in minimally invasive interventions where the CM is operating in confined cluttered spaces near sensitive organs. Future work includes extension of this work to localizing contact point, exploring other machine learning algorithms, as well as the analysis and study of the generalization capability of the proposed method to other CMs and sensors.

%\bibliographystyle{splncs04}
%\bibliography{main}

% The following has been inserted from the .bbl file (downloaded from the logs file within the overleaf project)

%
\end{document}